\documentclass[]{bytedance_seed}

\usepackage[toc,page,header]{appendix}

\usepackage{minitoc}

\usepackage{threeparttable}
\usepackage{colortbl} 
\usepackage{booktabs} 
\usepackage{amsmath}
\usepackage{siunitx} 
\usepackage[utf8]{inputenc} 
\usepackage[T1]{fontenc}    
\usepackage{url}            
\usepackage{booktabs}       
\usepackage{amsfonts}       
\usepackage{nicefrac}       
\usepackage{booktabs} 
\usepackage{multirow}
\usepackage{wrapfig}
\usepackage{amssymb}
\usepackage{epigraph}
\usepackage{makecell}
\usepackage{listings}
\usepackage{pdflscape}
\usepackage[labelfont=bf]{caption} 

\usepackage[ruled,vlined]{algorithm2e}
\usepackage{color, soul}
\usepackage{microtype}      
\usepackage{xcolor}         
\usepackage{geometry}
\usepackage{times}
\usepackage{ragged2e}
\usepackage{amsmath}
\usepackage{bm}
\usepackage{latexsym}
\usepackage{graphicx}
\usepackage{float}
\usepackage{longtable}
\usepackage{booktabs} 
\usepackage{multirow}
\usepackage{amssymb}
\usepackage{longtable}
\usepackage{tabu}
\usepackage{bbding}
\usepackage{awesomebox}
\usepackage{bbding}

\usepackage{geometry}
\usepackage{multirow}
\usepackage{tabularx}
\usepackage{makecell}
\usepackage{xcolor}
\usepackage{helvet}  
\usepackage{lmodern} 
\newcommand\ourmodel{ReTool\xspace}

\usepackage{roboto}

\title{ ReTool: Reinforcement Learning for Strategic Tool Use in LLMs  }

\author[*]{Jiazhan Feng}
\author[*]{Shijue Huang}
\author[]{Xingwei Qu}
\author[]{Ge Zhang}
\author[]{Yujia Qin}
\author[]{Baoquan Zhong}
\author[]{Chengquan Jiang}
\author[]{Jinxin Chi}
\author[\dagger]{Wanjun Zhong}

\affiliation[]{ByteDance Seed}

\contribution[*]{Co-first authors}
\contribution[\dagger]{Corresponding author}

\abstract{While reasoning models (e.g., DeepSeek R1) trained with reinforcement learning (RL), excel in textual reasoning, they struggle in scenarios requiring structured problem-solving, such as geometric reasoning, concise computation, or complex equation solving—areas where computational tools like code interpreters (CI) demonstrate distinct advantages. 
To bridge this gap, we propose \textbf{ReTool}, which enhances long-form reasoning with tool-integrated learning, including two key features: (1) dynamic interleaving of real-time code execution within natural language reasoning processes, and (2) an automated RL paradigm that allows policy rollouts with multi-turn real-time code execution and teaches the model in learning when and how to invoke tools based on outcome feedback.
\ourmodel employs a systematic training framework, beginning with synthetic cold-start data generation to produce code-augmented long-form reasoning traces for fine-tuning base models. Subsequent RL training leverages task outcomes as rewards to iteratively refine the model's tool use strategy, enabling autonomous discovery of optimal tool invocation patterns without human priors. 
Experiments on the challenging MATH Olympiad benchmark AIME demonstrate \ourmodel's superiority: Our 32B model achieves 67\% accuracy with 400 training steps, outperforming text-based RL baseline (40\% accuracy, 1080 steps) in efficiency and performance. Remarkably, \ourmodel-32B attains 72.5\% accuracy in extended settings, surpassing OpenAI's o1-preview by 27.9\%.
Further analysis reveals emergent behaviors such as code self-correction, signaling an ``aha moment'' in which the model autonomously masters adaptive tool use. These findings highlight the promise of outcome-driven tool integration for advancing complex mathematical reasoning and offer new insights into hybrid neuro-symbolic systems.}
\date{April 15, 2025}
\checkdata[Project Page]{\url{https://retool-rl.github.io/}}

\begin{document}

\maketitle
\begin{figure}[H]
    \centering
    \includegraphics[width=0.92\linewidth,]{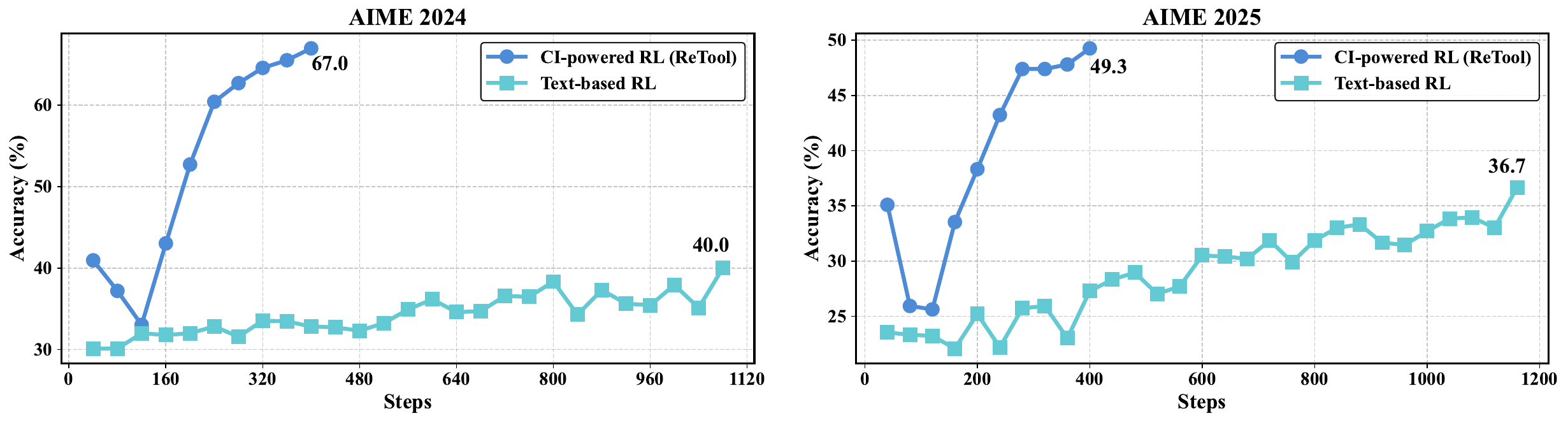}
    \vspace{-9pt}
    \caption{\small{AIME 2024 \& 2025 scores of \ourmodel and text-based RL baseline on the Qwen2.5-32B-Instruct model. 
    }}
    \label{fig:main-results}
\end{figure}

\section{Introduction}
Reinforcement learning (RL) has recently become a popular paradigm for enhancing the reasoning capabilities of large language models (LLMs), enabling them to explore and refine long chains of thought (CoT)~\cite{NEURIPS2022_9d560961,10.5555/3666122.3666639,luong2024reftreasoningreinforcedfinetuning,zhang2024mapneohighlycapabletransparent}. Reasoning models such as OpenAI o1~\cite{openai2024openaio1card} and DeepSeek R1~\cite{deepseekai2025deepseekr1incentivizingreasoningcapability} demonstrate strong performance in pure text-based reasoning tasks by learning to self-correct and engage in more deliberate, analytical thinking~\cite{Claude3.7,qwq32b,kimiteam2025kimik15scalingreinforcement}. These advances suggest early signs of metacognitive control, where models not only reason, but also monitor and revise their reasoning process.

Despite these advances, reasoning LLMs equipped with long chains of textual reasoning processes \cite{ouyang2022traininglanguagemodelsfollow} still show notable limitations in tasks that require precise numerical calculation or symbolic manipulation, such as geometric reasoning, precise computation, or complex equation solving.
In contrast, computational tools, such as code interpreters  (CI), can empower models with symbolic computation capabilities that go far beyond pure text-based reasoning. 
Unlike textual CoT~\cite{wei2023chainofthoughtpromptingelicitsreasoning} methods that rely solely on internal language patterns, code interpreters provide a formal and executable interface for enumeration, verification, and precise computation. 
This not only enables exact numeric validation of intermediate steps—dramatically reducing the ambiguity and compounding error often seen in textual reasoning~\citep{chen2023programthoughtspromptingdisentangling,wang2023mathcoderseamlesscodeintegration}, but also allows models to expand their solution search space via programmable exploration.

Recent works have explored prompting and supervised fine-tuning methods~\cite{Slow_Thinking_with_LLMs_3_Tool,pan-etal-2023-logic} to equip LLMs with tool-use capabilities. However, these approaches are limited to imitating the specifically-curated data distribution, often failing to generalize beyond seen patterns or adaptively decide when and how to invoke external tools. As a result, models may misuse tools or fall back on brittle heuristics that are not robust across diverse problem settings. To overcome these limitations, RL offers a principled solution: it enables models to explore flexible reasoning trajectories and learn tool-use strategies guided by outcome-based feedback. This paradigm not only incentivizes correct solutions, but also allows the model to discover nuanced behavioral patterns—such as how to recover from tool execution mistakes via self-correction, 
decide when to effectively invoke tool execution during the long-chain reasoning process.

In this work, we embrace the RL paradigm and introduce \textbf{\ourmodel}, a \textbf{Tool}-augmented \textbf{Re}inforcement learning framework explicitly designed to guide LLMs towards optimal strategies for leveraging external computational tools during reasoning. \ourmodel consists of two key components: First, we develop a data construction pipeline to curate a high-quality cold-start dataset that explicitly demonstrates when and how to invoke the code interpreter. 
This teaches the model an initial competency in tool usage and execution result analysis. 
Then, we apply tool-enhanced reinforcement learning to train the model in discovering the optimal tool manipulation reasoning strategy and adjusting its behavior through outcome-based rewards, going beyond what can be captured by supervised learning alone. During long-chain reasoning, the policy model rolls out by flexibly writing code blocks and achieving real-time execution results from a sandbox-style code interpreter to assist subsequent thinking.

We evaluate \ourmodel on the challenging MATH Olympiad benchmarks AIME2024 and AIME2025. Building on Qwen2.5-32B-Instruct~\cite{qwen2.5}, our model achieves 67.0\% accuracy on AIME2024 with only 400 training steps, significantly outperforming the text-based RL baseline, which achieves 40.0\% accuracy with 1080 training steps.
These substantial gains highlight that explicitly modeling tool-use as part of the decision process not only pushes the limits of model reasoning but also enhances training efficiency.
Furthermore, when trained on DeepSeek-R1-Distill-Qwen-32B~\cite{deepseekai2025deepseekr1incentivizingreasoningcapability}, our model demonstrates further improvements, surpassing competitive baselines such as QwQ-32B-Preview~\cite{qwq32b}, s1-32B~\cite{muennighoff2025s1simpletesttimescaling}, and OpenAI o1-preview~\cite{openai2024learning}.
This suggests that the RL training process inspires more efficient problem-solving strategies.
Additionally, our cold-start model based on Qwen2.5-32B-Instruct achieves an accuracy of 40.9\% on AIME2024, comparable to the text-based RL baseline based on same backbone (40.0\%), and significantly surpasses the non-trained Qwen2.5-32B-Instruct (26.7\%). These results demonstrate that our curated dataset effectively captures tool usage patterns within executable reasoning traces, and that CI-integrated training positively contributes to reasoning performance.

We further conduct a comprehensive analysis of CI cognitive behavior through RL training and identify several key findings. Our model demonstrates enhanced code utilization capabilities, enabling it to employ more accurate and complex code snippets; It also learns to invoke tools appropriately, select tool adaptively, structure tool calls effectively, and iteratively refine reasoning through emergent code self-correction capabilities.

\noindent\textbf{Our main contributions are summarized as follows:}

\begin{enumerate}
    \item We propose \textbf{\ourmodel}, a novel reinforcement learning framework that integrates code interpreter execution into the reasoning loop of LLMs. To equip the model with foundational capabilities for invoking the code interpreter, we curate a high-quality cold-start dataset through our developed pipeline. Furthermore, we design a reinforcement learning framework that supports interleaved code execution during rollout, enabling the model to iteratively explore, refine, and optimize its reasoning strategies through tool-augmented interactions guided by feedback from a sandboxed code interpreter.

    \item As shown in \cref{sec:insights},
    we conduct comprehensive empirical and behavioral analyses, and observe several key findings:
    (1) After RL training, the response length is reduced by approximately 40\% compared to that prior to training, showcasing the potential reasoning token efficiency of CI-powered reasoning;
    (2) During RL training, the code ratio, code lines and correct code counts show increase trends, and the code invocation timing becoming shifts earlier, indicating the improved code use capabilities and strategic tool usage development;
    (3) Emergent behaviors like code self-correction and adaptive tool selection can be observed during RL phase, bringing more advanced tool-augmented reasoning patterns.
    
\end{enumerate}

\section{Methodology} 
In this section, we introduce \ourmodel, a CI-powered RL framework designed to address math problem-solving tasks. We begin with an overview of \ourmodel. Next, we describe our cold-start training, including the data construction pipeline and supervised fine-tuning (\cref{sec:SFT}). We then outline our reinforcement learning pipeline, enhanced by a code interpreter sandbox, to further enhance strategic tool usage development (\cref{sec:RL}).

\subsection{Overview} \label{sec}
Our methodology consists of two primary stages: cold-start supervised fine-tuning followed by reinforcement learning with interleaved code execution rollout.
Firstly, we collect data through our designed pipeline for cold-start supervised fine-tuning (SFT), which provides a robust initialization for the reinforcement learning phase. 
To enhance our model's tool utilization capabilities, we introduce a specialized tool-using reinforcement learning pipeline that enhances the model's ability to appropriately select and apply tools during the reasoning process.

\subsection{Cold-start for Tool-Integrated Reasoning Foundation} \label{sec:SFT}
We designed a pipeline for collecting and curating high-quality data. Specifically, we begin by gathering existing mathematical reasoning data from diverse sources, including open-source datasets such as Open-Thoughts~\citep{openthoughts}. Subsequently, we implement a dual-verification approach combining human expert curation and Deepseek-R1~\cite{deepseekai2025deepseekr1incentivizingreasoningcapability} evaluation to filter invalid data. 
Through these steps, we collect a high-quality text-based reasoning dataset, denoted as $\mathcal{D}_\text{init}$.

\begin{figure}
    \centering    \includegraphics[width=0.93\linewidth]{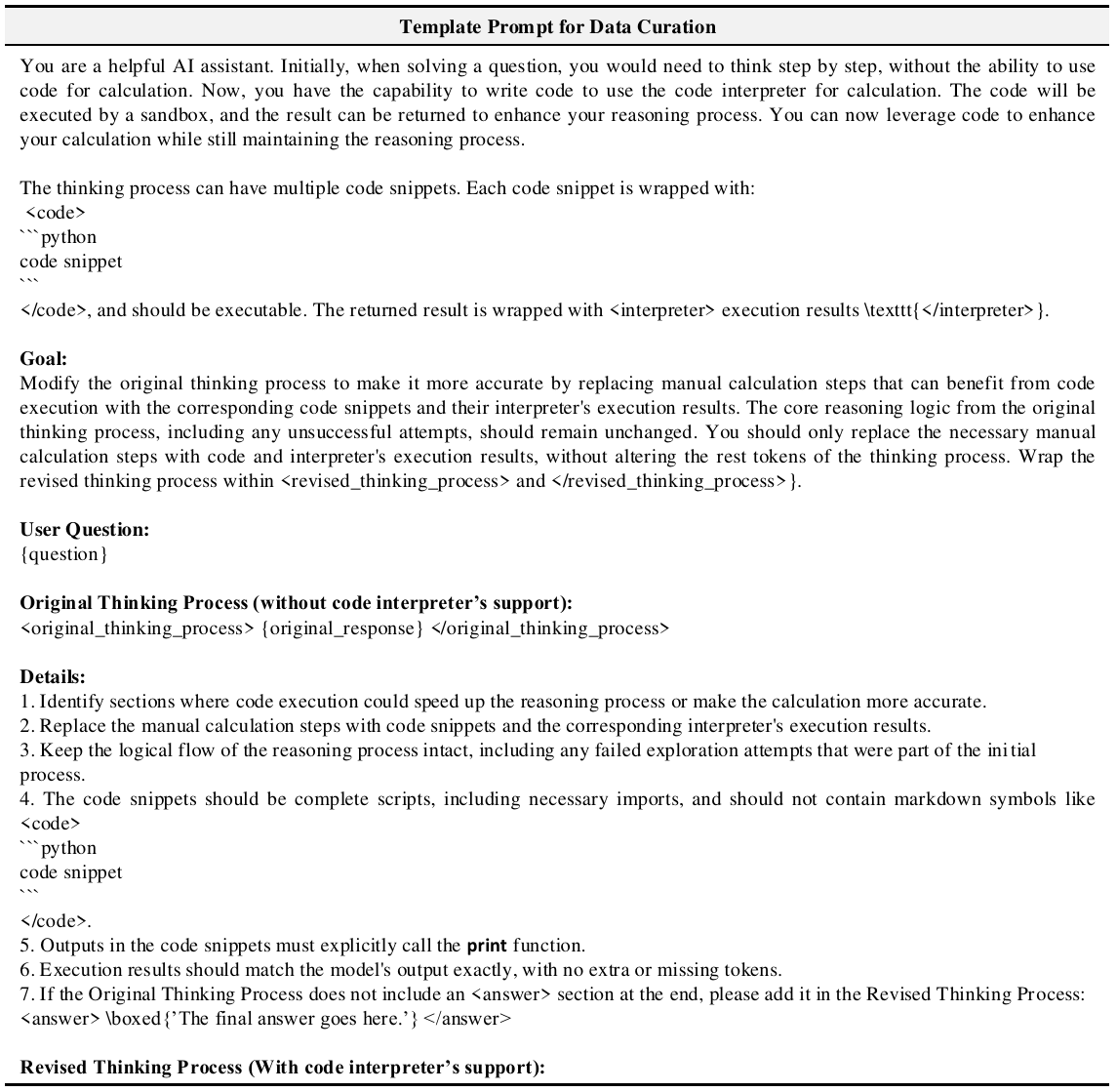}
    \caption{Template Prompt for Data Curation.}
    \label{fig:data-template}
\end{figure}

Based on $\mathcal{D}_\text{init}$, we further construct code-integrated reasoning data in an automatic manner.  
We first utilize a structured prompt template (detailed in \figurename~\ref{fig:data-template}) for transformation, which modifies the original thinking process by replacing manual calculation steps that can benefit from code execution with the corresponding code snippets and their interpreter's execution results.
Following this initial transformation, we apply a two-stage verification protocol. The first stage focuses on format verification, which improves readability and ensures consistent syntax that that enables the efficient detection of computational tool invocation triggers during subsequent reinforcement learning phases. The second stage entails answer verification, where we eliminate data samples whose final outputs do not align with the correct solutions to the mathematical problems. 
Finally, we collect a dataset $\mathcal{D}_\text{CI}$ that consist of code-augmented long-form reasoning traces.

\ourmodel employs supervised fine-tuning to learn when and how to invoke the code interpreter from the aforementioned dataset $\mathcal{D}_\text{CI}$, thereby enhancing the model's capability to appropriately utilize computational tools.

\begin{figure}[t]
    \centering    \includegraphics[width=0.9\linewidth]{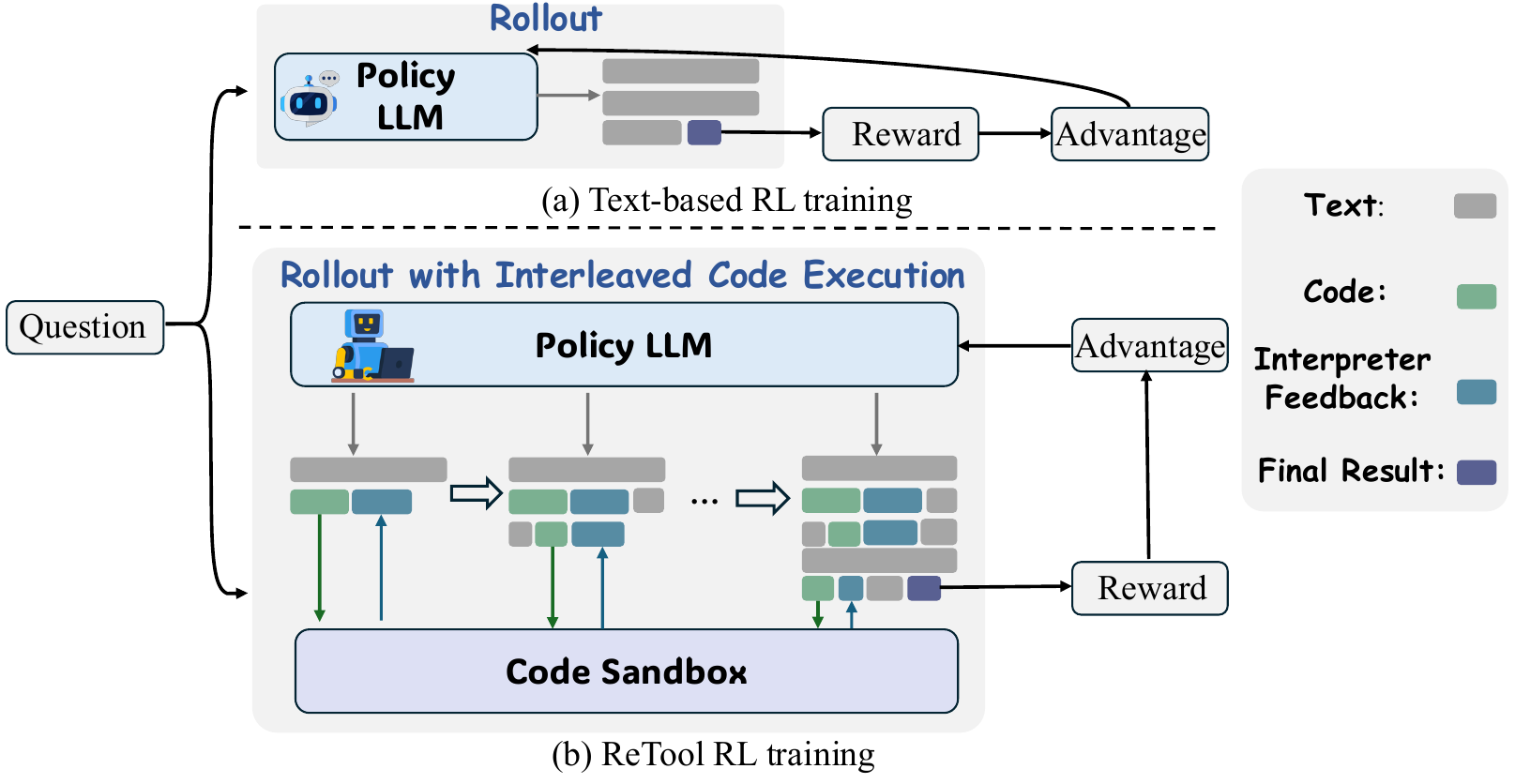}
    \caption{\small{
    Demonstration of text-based RL training process and \ourmodel's RL training process.}}
    \label{fig:model_rl}
\end{figure}

\subsection{\ourmodel: Reinforcement Learning for Strategic Tool Use} \label{sec:RL}

\subsubsection{Training Algorithm}
We train \ourmodel based on PPO algorithm~\cite{schulman2017proximalpolicyoptimizationalgorithms}, it updates policy with the following objective:
\begin{small}
    \begin{equation}
\begin{aligned}
\mathcal{J}_\text{PPO}(\theta) = \mathbb{E}_{(q,a)\sim \mathcal{D},o_{\le t}\sim\pi_{\theta_{\text{old}}}(\cdot\mid q)}
\Bigg[ 
\min \Bigg( \frac{\pi_{\theta}(o_t\mid q,o_{<t};\mathcal{CI})}{\pi_{\theta_{\text{old}}}(o_t\mid q,o_{<t};\mathcal{CI})} \hat{A}_t,  
\ \text{clip} \Bigg( \frac{\pi_{\theta}(o_t\mid q,o_{<t};\mathcal{CI})}{\pi_{\theta_{\text{old}}}(o_t\mid q,o_{<t};\mathcal{CI})}, 1 - \varepsilon, 1 + \varepsilon \Bigg) \hat{A}_t \Bigg) \Bigg],
\label{eq:ppoloss}
\end{aligned}
\end{equation}
\end{small}where $\pi_{\theta}$ is policy model, $\pi_{\theta_{\text{old}}}$ is reference model, $\pi_{\theta}(o_t\mid q,o_{<t};\mathcal{CI})$ represents the rollouts with interleaved code execution and feedback from code interpreter.
 
We modify PPO to better adopt tool integrated reasoning. During training, the policy LLM will collaborate with a code sandbox to generate rollouts with multi-turn real-time code
execution for solving given problems. We implement a rule-based outcome reward to enable the model with the flexibility to autonomously explore and develop strategies for code usage awareness, code selection, timing of code invocation, and further diverse behaviors. 

\textbf{Reward Design}
To teach the model in learning when and how to invoke tools, we implement a rule-based accuracy reward to optimize the model. 
The accuracy reward evaluates response correctness. 
We require the model to present final answers in a specified format (e.g., within \texttt{\textbackslash boxed\{\}}), enabling reliable rule-based verification. The reward is formulated as:
\begin{small}
    \begin{equation}
    R( a, \hat{a}) = 
    \begin{cases} 
    1, & \texttt{is\_equivalent}(a, \hat{a}) \\
    -1, & \text{otherwise} 
    \end{cases}
\end{equation}
\end{small}where $a$ and $\hat{a}$ represent the ground-truth answer and the predicted answer, respectively.
We simplify the reward design aim to alleviate reward hacking and promote more diverse problem-solving behaviors based on mere outcome feedback without considering code executability reward.

\begin{figure}[h]
    \centering    \includegraphics[width=0.73\linewidth]{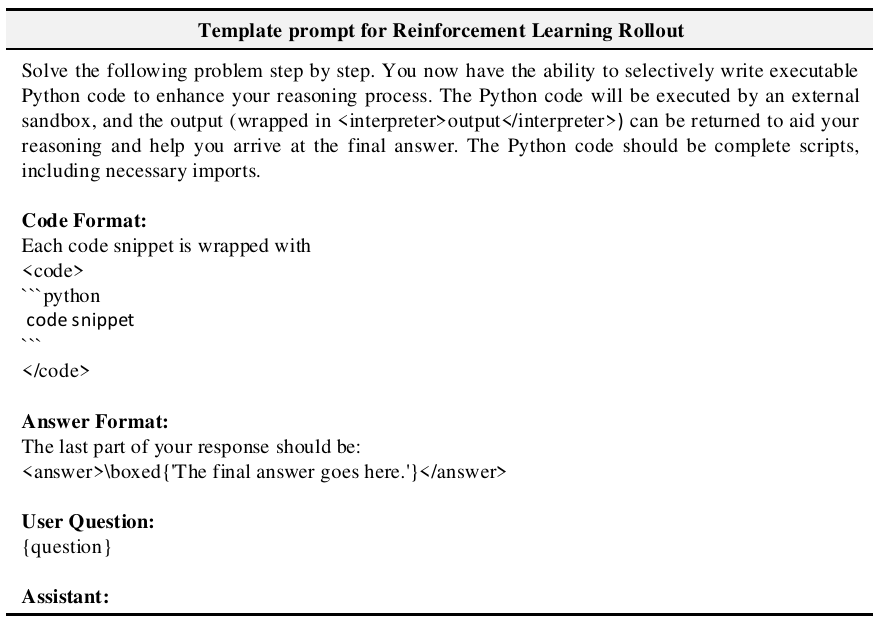}
    \caption{Template prompt for \ourmodel rollout.}
    \label{fig:rotool-template}
\end{figure}

\textbf{Rollout with Interleaved Code Execution}
To facilitate the integration of reasoning and executable code within the model, we propose a rollout approach that dynamically supports interleaved real-time code execution with natural language reasoning processes. 
As depicted in \figurename~\ref{fig:model_rl} (b), our rollout process differs from the conventional approach, which typically generates only text-based reasoning (as shown in \figurename~\ref{fig:model_rl} (a)). By contrast, our rollout approach integrates the collaboration of a policy LLM with an external code sandbox, enabling the production of hybrid content that combines text, code snippets, and real-time interpreter feedback.
Concretely, we utilize a prompt template (\figurename~\ref{fig:rotool-template}) to guide the model in interacting with the code sandbox by utilizing tags $\texttt{<code>}\texttt{</code>}$ to explicitly mark the boundaries of generated codes. 
During the rollout process, policy model generate text-based reasoning $t_1$ 
when a code termination trigger ($\texttt{</code>}$) is detected, the generation pause and the generated code $c_1$  is parsed and send to code sandbox environment for execution.
Upon completion, the sandbox's output $f_1$  (successful results or error messages) is filled within $\texttt{<interpreter>} \texttt{</interpreter>}$ tags and fed back to the model, which continues generating the rollout until either providing a final answer $o$ or producing a new code snippet, ultimately producing a hybrid reasoning trajectory $[t_1 \oplus c_1 \oplus f_1 \oplus ... \oplus o]$.

Notably, our approach returns both successful code execution results and interpreter error messages to the model. This dynamic feedback mechanism enables the model to iteratively explore, refine, and optimize its reasoning and tool usage strategies.

\subsubsection{Training Details}

\textbf{Cold-start \& RL }
For training, we employ the VeRL framework\footnote{\url{https://github.com/volcengine/verl}}. We adopt PPO as our RL method. We train our model on curated cold-start data for two epochs. Regarding hyperparameters, we utilize the AdamW optimizer with an initial learning rate of 1e-6. We define the expected maximum sequence length as 16384 tokens. For training, the mini-batch size is set to 512, and the KL coefficient is set to 0.0. We use Qwen2.5-32B-Instruct~\citep{qwen2025qwen25technicalreport} as the main backbone.

\textbf{Interpreter Feedback Mask.} We mask out the $\texttt{<interpreter>} \texttt{</interpreter>}$ feedback output from the loss computation. 
This sandbox-based output masking approach blocks external tokens from interfering with loss calculations, ensuring training stability and preserving the model's inherently generated coherent reasoning sequences from disruption.

\textbf{KV-Cache Reuse.} In order to reduce the memory cost during rollout, when each time the code termination trigger ($\texttt{</code>}$) is detected, we will cache all the KV-cache before code execution and only calculate and append the KV-cache from the interpreter feedback ($\texttt{<interpreter>} \texttt{</interpreter>}$). This will largely reduce the KV-cache for each rollout.

\textbf{Sandbox Construction.} To accelerate the RL training process, we design a asynchornous code sandbox environment. The sandbox pods function as workers in a pool, independently pulling tasks based on their current capacity, creating an efficient load-balancing mechanism. This distributed asynchronous approach accelerates RL training by enabling parallel environment interactions across multiple threads,
It prevents slower threads from creating bottlenecks and ensures optimal resource utilization, maintaining continuous throughput during the training process.

\section{Experiment}
In this section, we evaluate the performance of \ourmodel, and conduct comprehensive analysis on the behavior of model outputs.

\subsection{Evaluation Setup}
To ensure a stable evaluation, we repeat the evaluation set AIME2024\&2025 32 times and report the overall average accuracy to estimate pass@1. The inference hyperparameters of evaluation are set to temperature 1.0 and top-p 0.7.
We compare \ourmodel with competitive baselines, including Qwen2.5-Math-72B-Instruct~\cite{yang2024qwen25mathtechnicalreportmathematical}, 
Qwen2.5-Math-72B-Instruct-TIR~\cite{yang2024qwen25mathtechnicalreportmathematical}, 
Sky-T1~\cite{sky_t1_2025},
DeepSeek-R1-Zero-Qwen-32B~\cite{deepseekai2025deepseekr1incentivizingreasoningcapability}, 
QwQ-32B-Preview~\cite{qwq32b},
s1-32B~\cite{muennighoff2025s1simpletesttimescaling},
OpenAI o1-preview~\cite{openai2024learning}.
To verify the effectiveness of our \ourmodel, we also compare the performance with RL without tool-using, i.e. Text-based RL (Qwen2.5-32B-Instruct).
And for the results of baselines, we report the avg@k by coping from corresponding literature source as pass@1.

\definecolor{mygray}{gray}{0.92}
\begin{table}[ht]
\centering
\resizebox{0.75\linewidth}{!}{%
\begin{tabular}{l|cc}
\toprule[1.5pt]
Model & AIME2024 (pass@1) & AIME2025 (pass@1) \\
\midrule
\multicolumn{3}{c}{\textit{Existing Baselines}} \\
\midrule
Qwen2.5-Math-72B-Instruct & 30.0 & - \\
Qwen2.5-Math-72B-Instruct-TIR & 40.0 & - \\
Sky-T1 & 43.3 &  -\\
OpenAI o1-preview & 44.6 & 37.9 \\
DeepSeek-R1-Zero-Qwen-32B & 47.0 & - \\
QWQ-32B-Preview & 50.0 & 33.5 \\
s1-32B & 56.7 & - \\

\midrule
\multicolumn{3}{c}{\textit{CI-powered RL}} \\
\midrule
\rowcolor{mygray} \ourmodel (Qwen2.5-32B-Instruct) & 67.0 & 49.3 \\
\rowcolor{mygray} \ourmodel (DeepSeek-R1-Distill-Qwen-32B) & 72.5 & 54.3 \\

\midrule
\multicolumn{3}{c}{\textit{Ablations on Qwen2.5-32B-Instruct}} \\
\midrule
w/o Training (Base Model) & 26.7 & - \\
w/o CI (Text-based RL$^\spadesuit$) & 40.0 & 36.7 \\
w/o RL (only Cold-start$^\diamondsuit$) & 40.9 & 34.5 \\

\bottomrule[1.5pt]
\end{tabular}
}
\caption{Main results. $^\spadesuit$The Text-based RL method includes a text-based cold-start SFT initialization to ensure a fair comparison. $^\diamondsuit$The inference process of the Cold-start model also incorporates code interpreter.}
\label{tab:main}
\end{table}

\subsection{Main Results}
As shown in Table~\ref{tab:main}, \ourmodel enables the LLM to flexibly leverage the code interpreter during the RL stage, leading to substantial performance improvements. Specifically, \ourmodel (Qwen2.5-32B-Instruct) achieves accuracies of 67.0\% on AIME2024 and 49.3\% on AIME2025 with only 400 training steps. This markedly outperforms the text-based RL baseline (Qwen2.5-32B-Instruct), which attains 40.0\% and 36.7\% on the respective benchmarks despite using over 1000 training steps. These findings indicate that the tool-integrated learning paradigm employed by \ourmodel not only enhances the model's reasoning capabilities but also improves training efficiency.
Furthermore, on AIME2024, \ourmodel (Qwen2.5-32B-Instruct) surpasses the competitive baseline s1-32B by 10.3\%. Similarly, on AIME2025, it achieves an 11.4\% gain over OpenAI's o1-preview. When combined with a more advanced backbone, \ourmodel (DeepSeek-R1-Distill-Qwen-32B) further improves performance, achieving scores of 72.5\% on AIME2024 and 54.3\% on AIME2025.
These results suggest that more effective problem-solving strategies are discovered during the RL training process.

Moreover, our cold-start model based on Qwen2.5-32B-Instruct achieves an accuracy of 40.9\% on AIME2024, closely aligning with the performance of the text-based RL baseline (40.0\%), and substantially surpassing the base model (26.7\%). These results demonstrate that our curated dataset effectively captures tool usage patterns within executable reasoning traces, and that CI-integrated training contributes positively to reasoning performance.

\subsection{Cognitive Analysis} \label{sec:insights}
We present a comprehensive analysis and highlight several key findings from our exploration, including:
(1) The dynamics of code interpreter (CI)-related behaviors throughout the RL process;
(2) The emergence of self-correcting capabilities;
(3) Differences in code purpose before and after RL;
(4) Distinctions between CI-powered reasoning and text-based reasoning.

\begin{figure}[ht]
    \centering
    \includegraphics[width=\linewidth]{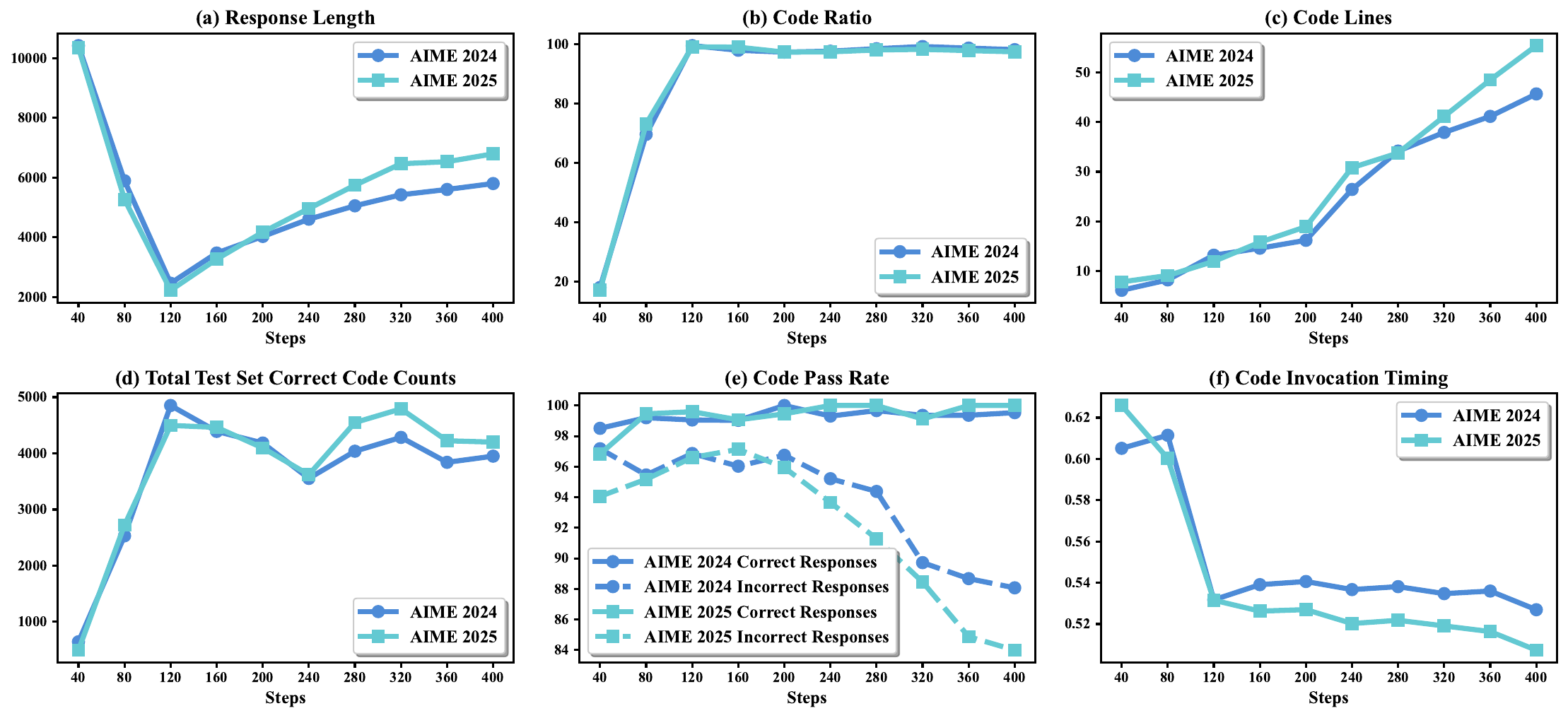}
    \caption{\small{CI-related behavior evolution during RL training.}}
    \label{fig:cog}
\end{figure}

\noindent \textbf{CI-related Behavior Evolution.}
To gain deeper insights into the RL process of \ourmodel, we systematically evaluated CI-related metrics. Specifically, we computed these metrics by analyzing model-generated outputs on the AIME2024 and AIME2025 datasets based on each saved checkpoint during RL training. The results are illustrated in \figurename~\ref{fig:cog}, and our analysis comprises:

\begin{itemize}
    \item \textbf{Response Length} (\figurename~\ref{fig:cog} (a)): 
    We calculated the average response length and observed a distinct trend: the generated response length initially declines sharply, later followed by a relatively gentle increase. We attribute the initial decline to the replacement of complex computational processes with more concise code, while the subsequent rise is likely due to the emergence of more diverse and complex code behaviors during RL training. Notably, the final average response length  remains 40\% shorter than that before RL training (i.e., from 10k to 6k). This suggests that the CI-powered reasoning approach potentially enhances efficiency of reasoning token utilization ratio by replacing intricate computational processes with code.

    \item \textbf{Code Ratio} (\figurename~\ref{fig:cog} (b)): The ratio of responses that contain code are also calculated. Analysis reveals that throughout the RL training process, the average code ratios exhibit a total upward trend and end with covering nearly 98\% percent of all questions. This suggests that the model’s proficiency in code utilization improved progressively during the RL process, facilitating strategic tool usage development.
    
    \item \textbf{Code Lines} (\figurename~\ref{fig:cog} (c)): The lines of generated code reflects its complexity to some extent. Observations show that the average code lines in responses exhibits a consistent upward trend throughout training. By the end of RL training, the final average code lines is nearly fivefold higher than that before RL training. This trend suggests that the model has learned more complex code strategies during the RL phase.

    \item \textbf{Total Test Set Correct Code Counts} (\figurename~\ref{fig:cog} (d)): 
    The number of total correct code counts on test set exhibits an overall upward trend during RL training, increasing from 1k to 5k. This improvement indicates the enhanced proficiency in leveraging code tools.
    
    \item \textbf{Code Pass Rate} (\figurename~\ref{fig:cog} (e)): 
    The CI-powered reasoning process involves generating intermediate code that may initially be incorrect, followed by iterative refinement based on interpreter feedback to produce executable code, so we report the average pass rate of last code in incorrect responses.
    Our analysis reveals that the code pass rate for correct responses remains consistently high, approaching 100\%, while the code pass rate for incorrect responses exhibits a declining trend. This pattern suggests that code executability impacts the reasoning process and final result. 
    
    \item \textbf{Code Invocation Timing} (\figurename~\ref{fig:cog} (f)): 
    We also calculate the code invocation timing, which is determined by dividing the start position of code by the total length of the response. This metric reflects the timing of code invocation within the response. The results show that the code invocation timing advances during the RL training process, indicating that the model learns to determine the timing for tool usage.
\end{itemize}

\begin{figure}[ht]
    \centering
    \includegraphics[width=0.8\linewidth]{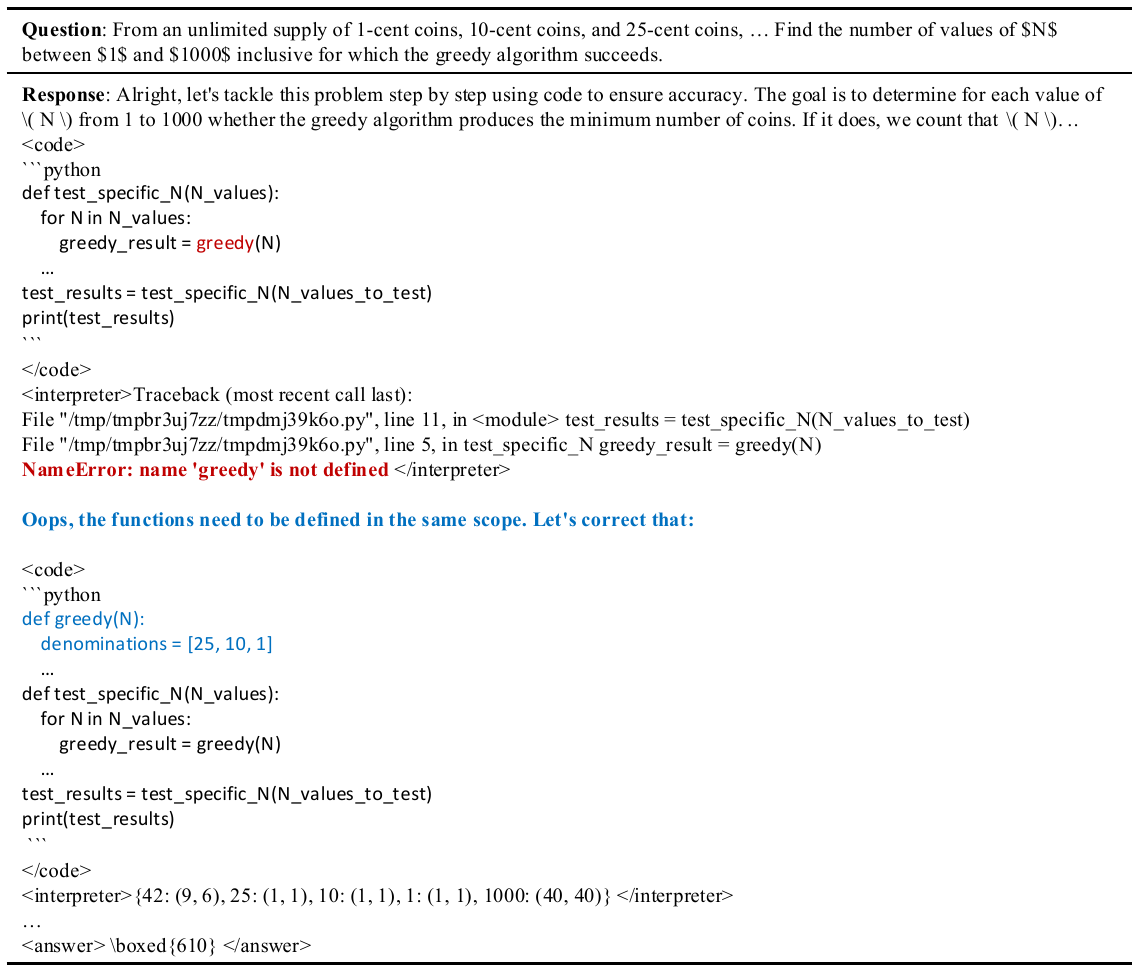}
    \caption{\small{The case of ``aha moment'' about code self-correction.}}
    \label{fig:aha}
\end{figure}

\noindent \textbf{``Aha Moment'' of Code Self-correction.}
Interestingly, our model exhibits an emergent ability to self-correct non-executable code, despite the absence of explicit training data for code self-correction. As shown in Figure \ref{fig:aha}, the model initially produced code that failed to execute due to the undefined function ``greedy()''. Upon receiving feedback from the interpreter, the model recognized the error and responded with the reflection: ``\textbf{Oops, the functions need to be defined in the same scope. Let's correct that}.'' It then proceeded to generate a revised, executable version of the code that included all necessary function definitions. This emergent behavior suggests that reinforcement learning can foster metacognitive capabilities, enabling the model to iteratively refine its generated code to address more complex problems.

\begin{figure}[h]
    \centering
    \includegraphics[width=0.65\linewidth]{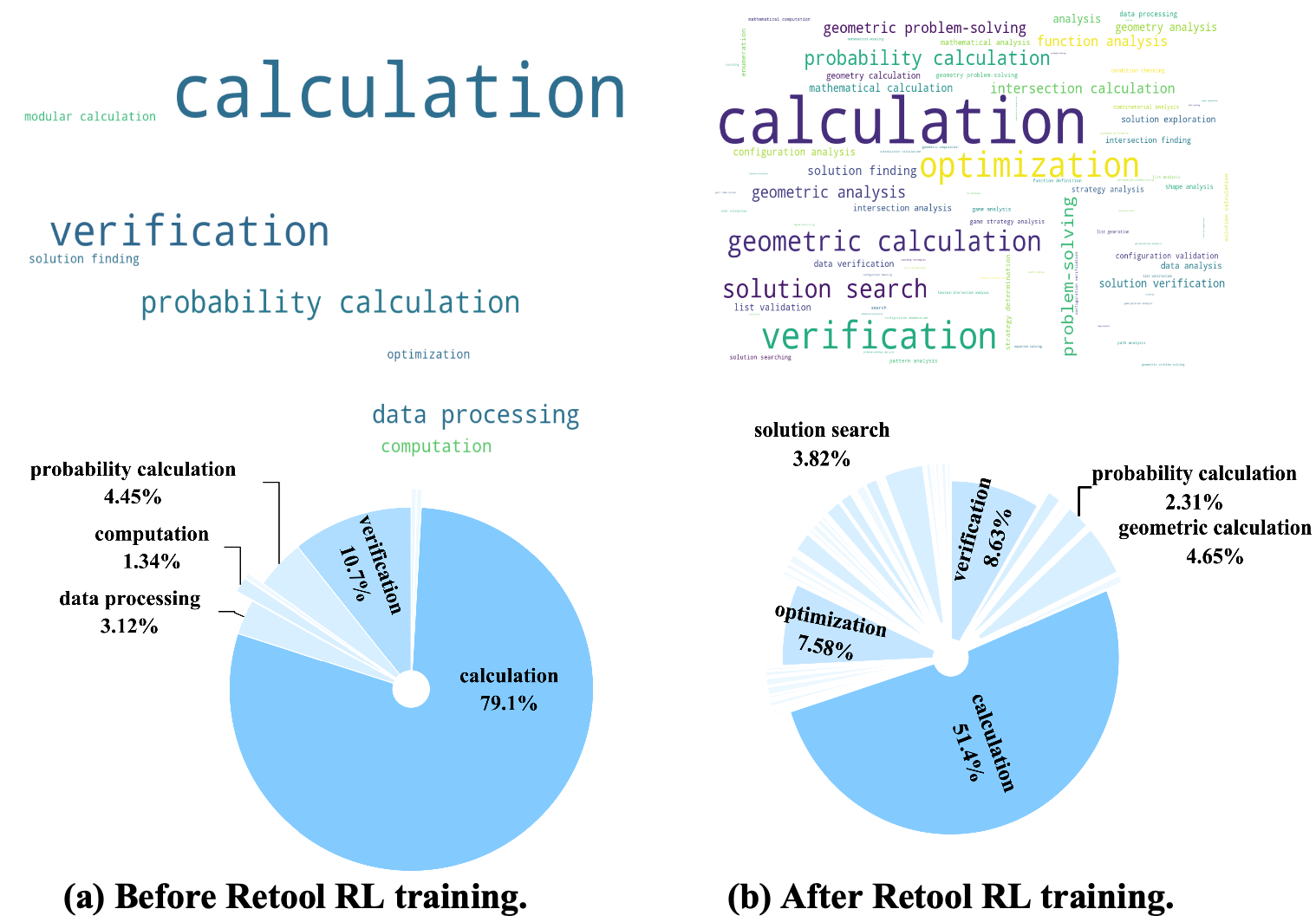}
    \caption{\small{Code purpose analysis.}}
    \label{fig:purpose}
\end{figure}

\noindent \textbf{Code Purpose Analysis.}
We also analysis the differences in code purposes before and after RL training, which reflects the types of code. We employ Doubao-1.5-pro\footnote{\url{https://team.doubao.com/zh/special/doubao_1_5_pro}} to classify the primary purpose of code snippets based on their contextual information, then compute the frequency of code purposes that appear more than once, and the results are depicted in Figure \ref{fig:purpose}. 
The word clouds reveal that calculation and verification are the dominant purposes of code in CI-powered reasoning. After RL training, the code purposes in our model become more diverse, which demonstrates the metacognitive development of adaptive tool selection and enhances the generalizability of \ourmodel to a broader range of problems.

\begin{figure}[ht]
    \centering
    \includegraphics[width=\linewidth]{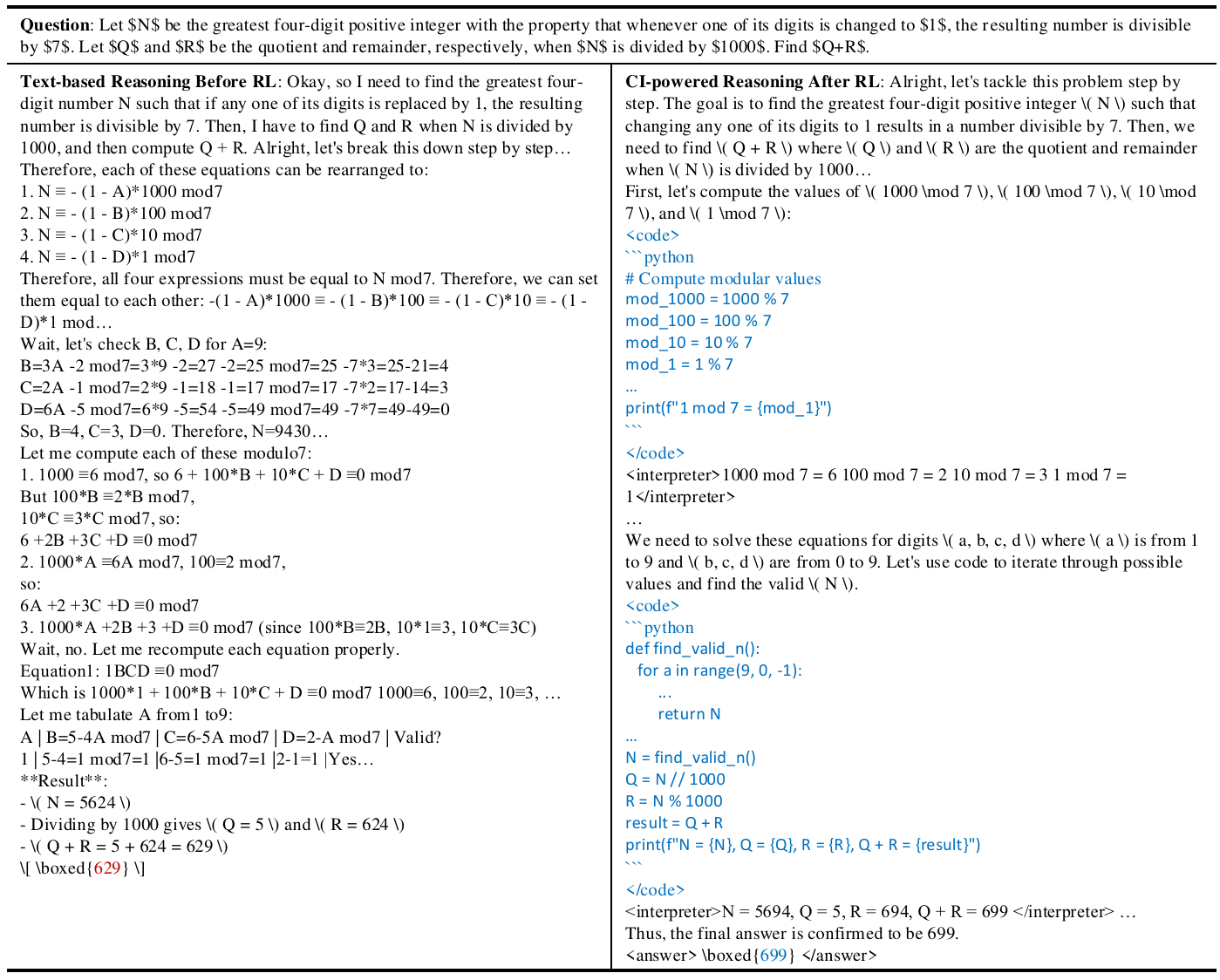}
    \caption{\small{Case of CI-powered Reasoning vs. Text-based Reasoning.}}
    \label{fig:case1}
\end{figure}

\noindent \textbf{CI-powered Reasoning vs. Text-based Reasoning.}
We present a case study to illustrate the distinction between CI-powered reasoning after reinforcement learning (RL) training and conventional text-based reasoning prior to RL training, as illustrated in Figure \ref{fig:case1}. When faced with the same question, text-based reasoning relies on a ``laborious'' text-only calculation process, which is prone to numerical errors and often results in incorrect inference outcomes. In contrast, CI-powered reasoning substitutes this complex calculation process with concise code. This approach not only ensures computational accuracy through the assistance of an external code interpreter but also enables the model to focus more effectively on holistic reasoning strategies.

\section{Background and Related Work}

\subsection{LLM Reasoning}
Recent advancements in large language models (LLMs) \citep{NEURIPS2022_9d560961,10.5555/3666122.3666639,luong2024reftreasoningreinforcedfinetuning,openai2024openaio1card,kimiteam2025kimik15scalingreinforcement,deepseekai2025deepseekr1incentivizingreasoningcapability,xai2023grok,Claude3.7,team2023gemini,qwen2.5} indicate significant progress toward cognitive abilities similar to human metacognition through Chain-of-Thought (CoT) prompting. CoT prompting, first introduced by \citet{wei2023chainofthoughtpromptingelicitsreasoning}, enhances the reasoning capabilities of LLMs by leveraging step-by-step natural language descriptions, significantly improving performance on various reasoning tasks. Building upon this foundation, recent research has shifted focus from train-time scaling to test-time scaling \citep{snell2024scalingllmtesttimecompute}, where additional computational resources are allocated during inference to enable the generation of intermediate reasoning steps. Techniques such as stepwise preference optimization \citep{lai2024stepdpostepwisepreferenceoptimization}, Monte Carlo Tree Search (MCTS) \citep{xie2024montecarlotreesearch}, and reinforcement learning \citep{luong2024reftreasoningreinforcedfinetuning} have been employed to improve multi-step and long-form mathematical reasoning. Advanced models like OpenAI-o1 \citep{openai2024openaio1card} and DeepSeek-R1 \citep{deepseekai2025deepseekr1incentivizingreasoningcapability} exemplify the effectiveness of CoT-based reasoning. Complementing CoT, Program-of-Thought (PoT) reasoning, introduced by \citet{chen2023programthoughtspromptingdisentangling} and \citet{gao2023palprogramaidedlanguagemodels}, integrates external computational tools—such as Python interpreters—to simplify and validate complex reasoning steps, resulting in enhanced accuracy.

\subsection{Tool Integrated Reasoning}
Tool-integrated reasoning was first introduced to help LLMs solve computationally intensive mathematical problems with the integration of programming strategies~\cite{chen2023programthoughtspromptingdisentangling,yue2023mammothbuildingmathgeneralist, jin2025searchr1trainingllmsreason, song2025r1searcherincentivizingsearchcapability,10.1145/3626772.3661381}. Building on this foundation, \citet{wang2023mathcoderseamlesscodeintegration} proposed an iterative approach that combines textual reasoning with code execution to mutually verify and enhance reasoning accuracy.
More recently, \citet{Slow_Thinking_with_LLMs_3_Tool} integrated code execution into the reasoning process by performing supervised fine-tuning on self-curated code-integrated CoT data. However, this approach is inherently limited by its reliance on the specific data distribution, and cannot learn adaptive strategies for tool use—such as determining when and how to invoke tools—through reinforcement learning. 
A concurrent work~\citep{li2025torlscalingtoolintegratedrl} applied reinforcement learning to learn tool usage strategies on Qwen2.5-Math models~\cite{yang2024qwen25mathtechnicalreportmathematical} at 1.5B and 7B scales, but the performance remained suboptimal. We further scale up this line of research and propose \ourmodel, a framework that leverages reinforcement learning to strategically determine when and how to invoke the code interpreter. Our method outperforms Qwen-Math-72B-TIR~\cite{yang2024qwen25mathtechnicalreportmathematical} and o1-preview~\cite{openai2024learning} significantly on AIME2024 and AIME2025. 
We also present a comprehensive analysis of the learned tool-use behaviors and highlight several key findings regarding the model's cognitive patterns in code invocation after \ourmodel training.

\section{Conclusion}

In this paper, we propose ReTool, a novel reinforcement learning framework that empowers large language models to self-enhance their mathematical reasoning capabilities through effective Code Interpreter utilization. Our comprehensive experiments on AIME2024 and AIME2025 demonstrate that ReTool not only achieves superior accuracy compared to conventional text-based RL approaches, but also converges with significantly fewer training steps. Through careful data curation and our specialized tool-using pipeline, ReTool enables models to develop sophisticated computational intervention strategies, paving the way for more efficient and powerful tool-augmented reasoning in LLMs.

\section*{Acknowledgments}
We would like to thank Guang Shi,  Mingxuan Wang, Renjie Zheng, Chen Dun, and Yun Jiang for their support on this work.

\clearpage

\bibliographystyle{plainnat}
\bibliography{main}

\clearpage

\beginappendix

\section{Appendix}

\end{document}